%% file: main.tex
\documentclass[10pt,twocolumn,letterpaper]{article}

\usepackage{iccv}              

\usepackage{multirow}
\usepackage{booktabs}
\usepackage{amsmath}
\usepackage{placeins}
\usepackage[table]{xcolor}
\usepackage{amssymb}
\usepackage{float}
\usepackage{orcidlink}
\usepackage[accsupp]{axessibility}  


\definecolor{iccvblue}{rgb}{0.21,0.49,0.74}
\usepackage[pagebackref,breaklinks,colorlinks,allcolors=iccvblue]{}

\def\paperID{2323} 
\def\confName{ICCV}
\def\confYear{2025}

\title{Perspective-Aware Teaching: Adapting Knowledge for Heterogeneous Distillation}

\author{
  Jhe-Hao Lin\textsuperscript{1}, 
  Yi Yao\textsuperscript{1} \orcidlink{0000-0001-8227-5662}, 
  Chan-Feng Hsu\textsuperscript{1}, 
  Hong-Xia Xie\textsuperscript{2} \orcidlink{0000-0002-5652-4327},\\
  Hong-Han Shuai\textsuperscript{1} \orcidlink{0000-0003-2216-077X}, 
  Wen-Huang Cheng\textsuperscript{3} \orcidlink{0000-0002-4662-7875}\\
  \textsuperscript{1}National Yang Ming Chiao Tung University \quad
  \textsuperscript{2}Jilin University \quad
  \textsuperscript{3}National Taiwan University\\
  {\tt\small \{jimmylin0979.11, leo81005.ee10, cfhsu311510211.ee11, hhshuai\}@nycu.edu.tw}\\
  {\tt\small hongxiaxie@jlu.edu.cn, wenhuang@csie.ntu.edu.tw}
}

\begin{document}
\maketitle
\input{sec/0_abstract}    
\input{sec/1_intro}
\input{sec/2_related}
\input{sec/3_method}
\input{sec/4_exper}
\input{sec/5_ablation}
\input{sec/6_conclusion}
{
    \small
    \bibliographystyle{ieeenat_fullname}
    \bibliography{strings, main}
}


\end{document}

%% file: sec/0_abstract.tex
\begin{abstract}

Knowledge distillation (KD) involves transferring knowledge from a pre-trained heavy teacher model to a lighter student model, thereby reducing the inference cost while maintaining comparable effectiveness. Prior KD techniques typically assume homogeneity between the teacher and student models. However, as technology advances, a wide variety of architectures have emerged, ranging from initial Convolutional Neural Networks (CNNs) to Vision Transformers (ViTs), and Multi-Level Perceptrons (MLPs). Consequently, developing a universal KD framework compatible with any architecture has become an important research topic. In this paper, we introduce a perspective-aware teaching (PAT) KD framework to enable feature distillation across diverse architectures. Our framework comprises two key components. First, we design prompt tuning blocks that incorporate student feedback, allowing teacher features to adapt to the student model's learning process. Second, we propose region-aware attention to mitigate the view mismatch problem between heterogeneous architectures. By leveraging these two modules, effective distillation of intermediate features can be achieved across heterogeneous architectures. Extensive experiments on CIFAR, ImageNet, and COCO demonstrate the superiority of the proposed method. Our code is available at \url{https://github.com/jimmylin0979/PAT.git}.
\end{abstract}

%% file: sec/1_intro.tex
\begin{figure}
  \centering
  \centerline{\includegraphics[width=1.0\linewidth]{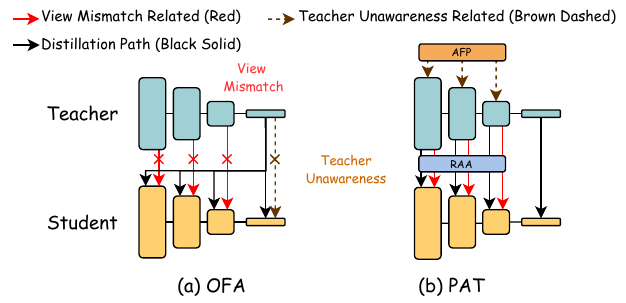}}
  \caption{High-level comparison between former state-of-the-art OFA and our method PAT. Due to the view mismatch (red lines) and teacher unawareness (brown dashed lines) problems, OFA can only use final logits to supervise student intermediate features (black solid lines), thereby restricting its utility in downstream tasks. In contrast, PAT enables feature imitation by solving these two issues with RAA and AFP modules respectively.}
  \label{fig_fig1}
  \vspace{-15pt}
\end{figure}

\section{Introduction}
\label{sec:intro}


Knowledge distillation (KD) has received increasing attention from both academic and industrial researchers in recent years. The goal is to develop a lightweight and efficient student model by transferring knowledge from an already trained, larger teacher model. This knowledge may come in the form of predictions \cite{zhao2022dkd, li2023curriculum}, or intermediate features \cite{chen2021reviewkd, chen2021cross, afd2021}, from the cumbersome pre-trained teacher model. Despite the significant progress in KD, there remains a gap in addressing cross-architecture distillation comprehensively. Existing KD approaches have predominantly focused on distillation between teacher and student models with homogeneous architectures, particularly feature-based approaches, leaving cross-architecture distillation relatively unexplored. For example, FitNet \cite{adriana2015fitnet} achieves good results in CNN-to-CNN format but performs poorly when distilling in a heterogeneous CNN-to-ViT scenario, only getting 24.06\% on CIFAR-100 with ConvNeXt-T teacher and Swin-P student model pair, which is lower than the basic logit-based method KD \cite{Hinton2015DistillingTK} by more than 50\%.



Heterogeneous KD, however, offers numerous advantages. First, it helps overcome architecture-specific limitations, allowing the student model to mitigate weaknesses inherent to individual architectures and achieve a better performance than learning from a homogeneous teacher model. For instance, CNNs can assimilate a more comprehensive understanding from ViTs, conversely, ViTs can augment their capacity for local feature extraction by leveraging insights from CNNs. Second, it enhances the robustness, as the student model learns from the varied error patterns and inductive biases of the teacher models, leading to better performance across different tasks and datasets. While a few studies have delved into the realm of heterogeneous architectures \cite{liu2022cross, zhao2023cross, ren2022coadvise}, it is noteworthy that these architectures often exhibit limitations in accommodating specific teacher-student pairings. As a result, a thorough restructuring of the link between the teacher and student models is essential whenever a new architecture is introduced into the system.

To overcome the existing limitations, OFA-KD \cite{hao2023ofa} is the first to propose a generic framework for heterogeneous KD by projecting the intermediate student features into the logits space and subsequently aligning them with the final predictions of the teacher model. The projection into a latent space with reduced architecture-specific details allows the distillation process to concentrate on high-semantic information, thereby enhancing distillation performance across different architectures. However, this projection results in a loss of spatial information that is crucial for downstream tasks. Furthermore, distilling based on the final high-semantic output of the teacher model may not fully exploit the valuable intermediate features.

To effectively leverage the richly informative intermediate layer features and facilitate distillation across diverse architectures, we revisit the fundamental feature-based distillation pipeline to identify the potential issues. As illustrated in Fig.~\ref{fig_fig1}, apart from the different perspectives inherent in different architectural biases, which we refer to as the view mismatch problem, teacher unawareness is a notable concern, whereby the teacher model fails to adjust based on the student's learning progress. Hence, a novel perspective-aware teaching (PAT) KD framework is proposed for distillation between heterogeneous architectures to tackle these issues. To address the view mismatch problem, a region-aware attention (RAA) module is proposed to let the student model learn how to reblend features from different patches and stages to achieve a similar perspective with the corresponding teacher features via the attention mechanism. Since the perspective is now similar, traditional stage-wise feature alignment across different architectures can be applied normally. For the teacher unawareness problem, an adaptive feedback prompt (AFP) module is introduced. The AFP module employs prompt tuning \cite{nie2023promptTuning} to generate distillation-friendly teacher features with additional branches. Moreover, the difference of the feature map between the student and teacher model, referred to as feedback, is integrated as extra input to the prompt tuning blocks, enabling the teacher model to dynamically adapt its features in response to the student model's learning trajectory. 





Our contributions are summarized as follows:
\begin{itemize}
    \item To address the view mismatch issue, we propose region-aware attention (RAA) to reblend student features to have a similar perspective with the teacher model. 
    \item To address the teacher unawareness problem, we propose an adaptive feedback prompt (AFP) that enables the teacher model to dynamically adapt its features in response to the student model's feedback.
    \item Extensive experiments manifest that PAT improves student model performance with gains of up to 16.94\% on CIFAR-100 and 3.35\% on ImageNet-1K and 3.50\% performance gain on COCO.
\end{itemize}

%% file: sec/2_related.tex
\section{Related Works}
\label{sec:related}

\subsection{Homogeneous Knowledge Distillation}

\noindent A majority of the KD methods \cite{zhao2022dkd, park2019rkd} predominantly concentrate on homogeneous architectures, particularly within feature-based approaches \cite{adriana2015fitnet, tian2019crd, wang2023improving}, as the intermediate features between such architectures exhibit greater similarity, thereby enhancing the efficiency of the distillation process. For instance, ReviewKD \cite{chen2021reviewkd} introduces a review mechanism that leverages multi-level information from the teacher model to direct the single-level learning of the student model, achieving commendable performance in the distillation process between CNN-based models. The aforementioned methods exhibit efficacy in distillation processes involving homogeneous architectures. Nonetheless, they fail to consider heterogeneous architectures. As such, they may encounter challenges like view mismatch and teacher unawareness when attempting to transfer knowledge from a heterogeneous teacher model.

\subsection{Generic Heterogeneous Knowledge Distillation}


As more model architectures are developed, creating a KD framework that works with any architecture has become crucial. Different architectures have unique biases \cite{raghu2021do}, which require careful design for KD and often limit it to specific teacher-student pairs \cite{liu2022cross, zheng2023distilling, ren2022coadvise, abnar2020transferring}. OFA-KD \cite{hao2023ofa} solves this by converting student features into a logits space, allowing alignment across architectures by removing architecture-specific details, but it loses important spatial information in features. In this work, we aim to design a feature-based distillation framework that keeps rich spatial information, improving both classification performance and downstream tasks.

\subsection{Adaptive Teacher in Knowledge Distillation} 




In typical KD approaches \cite{Hinton2015DistillingTK, chen2021reviewkd, mirzadeh2020takd, li2023curriculum}, the teacher model is first trained in isolation and then used to distill the knowledge to the student model. This results in static knowledge transfer that ignores the student’s learning dynamics, often leading to suboptimal outcomes. This phenomenon has also been validated in \cite{shayan2024effectiveTeacher}, where they attempted to improve student model learning by substituting the conventional teacher trained with cross-entropy loss with one trained using MSE loss. Although this approach yields better performance, training an entirely new teacher model often requires significantly higher computational resources. Another research direction draws inspiration from prompt tuning \cite{jia2022vpt}, exploring the possibility of leveraging task-specific prompts that can be adjusted to fine-tune the entire teacher model with minimal parameters \cite{kim2024promptkd, zhao2023cross}. For instance, CAKD \cite{zhao2023cross} integrates adaptable prompts into the teacher model, enabling it to manage distillation-specific knowledge while preserving its discriminative capability. In this paper, we propose using prompt tuning in a generic heterogeneous KD framework to create an adaptive teacher. Unlike previous methods that directly feed the teacher feature, we additionally incorporate student feedback to better align the adaptation with the student's learning process, further improving the performance.

%% file: sec/3_method.tex
\section{Methodology}
\label{sec:method}

\begin{figure*}[htb]
  \centering
  \centerline{\includegraphics[width=1.0\textwidth]{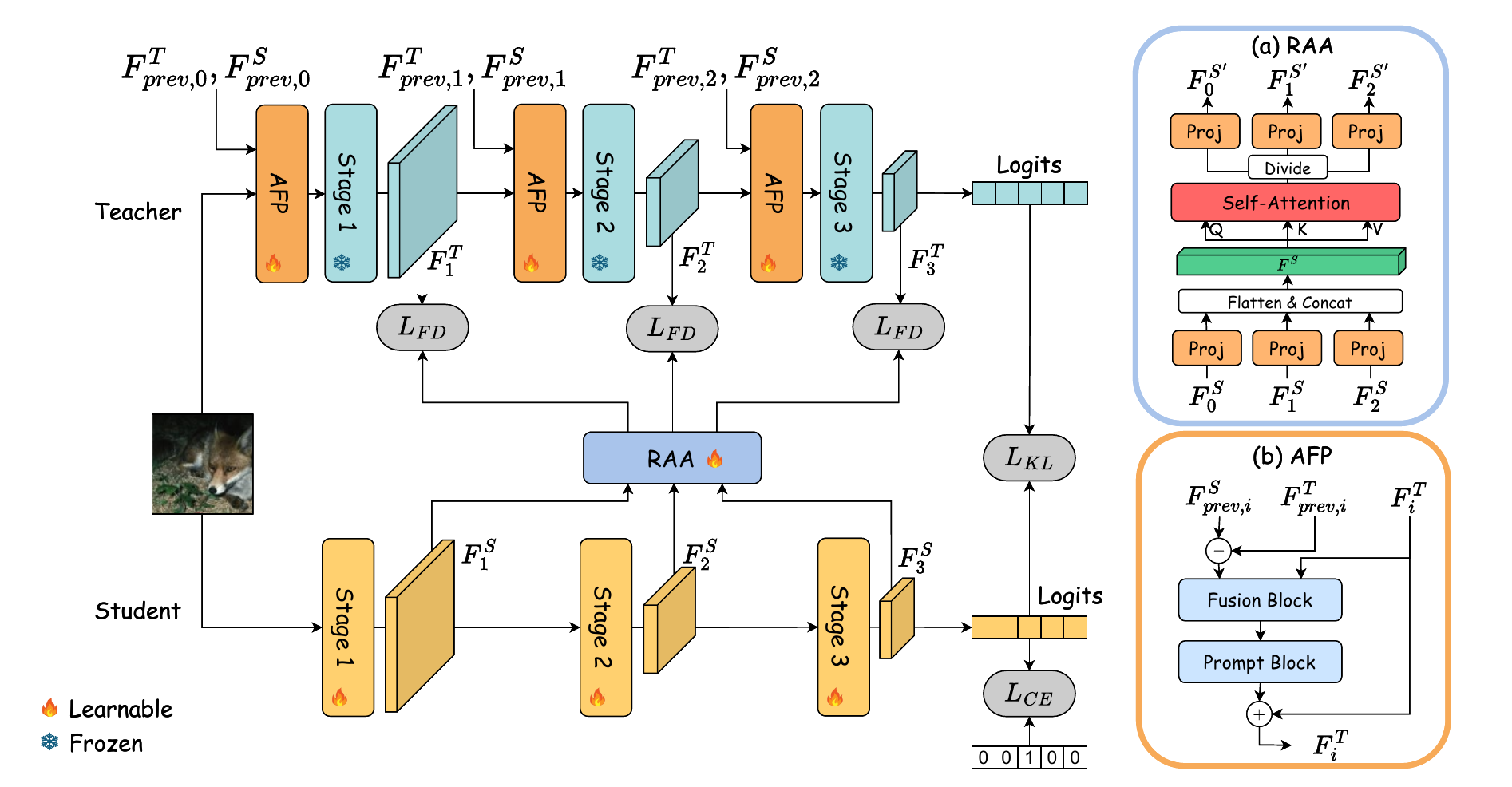}}
  \caption{A comprehensive depiction of the general structure of our PAT framework and proposed modules. (a) RAA: student features across all stages are concatenated and fed into an attention module to learn how to integrate a new feature with a perspective similar to that of the teacher model. (b) AFP: a prompt-tuning method is introduced to modify the output stage features of the teacher model with respect to the student model's learning process. \textit{Note}: Only
three stages are shown for convenience. In our experiment, all models are split into four stages.}
  \label{fig_framework}
  \vspace{-10pt}
\end{figure*}


In this section, we first review knowledge distillation (KD) and then introduce our PAT framework, which addresses existing issues such as view mismatch and teacher unawareness. We describe the adaptive feedback prompt (AFP) and region-aware attention (RAA) components, and summarize the objective function. An overview of the PAT framework is shown in Fig.~\ref{fig_framework}.

\subsection{Preliminaries}

Knowledge Distillation (KD) \cite{Hinton2015DistillingTK} aims to effectively transfer knowledge from the pre-trained teacher model to a more compact student model. Among the various types of KD, logits and features are the most commonly used forms of knowledge to transfer.

Logits-based distillation trains the student model to mimic the predictive distribution of the teacher model. This process can be formulated as:
\begin{equation}
    L_{LD} = L_{KL}(p^{T}, p^{S}),
\end{equation}
where $p^{T}$, $p^{S}$ represents the teacher and student model's confidence across all classes, respectively, and $L_{KL}$ denotes the KL-divergence function. For simplicity, we omit the temperature term here.

On the other hand, features-based distillation utilizes intermediate features from the teacher model to train the student model in a more fine-grained manner, which can be expressed as:
\begin{equation}
\label{equ_FD}
    L_{FD} = \Sigma^{4}_{i=1} D(F^{T}_{i}, M_{i}(F^{S}_{i})),
\end{equation}
where $F^{T}_{i}$, $F^{S}_{i}$ denote the features of the teacher and student model in the $i$-th stage, respectively. All teacher and student models are divided into four stages. $M_{i}$ represents the transformation applied to the student feature at the $i$-th stage, and $D$ denotes the distance function that measures the gap between the teacher and student features, which is typically the Mean Squared Error (MSE) function.

Overall, while logits-based distillation incurs minimal computational costs, its performance is generally inferior to that of the features-based approach. Despite this limitation, the architecture-agnostic nature of the predictive distribution in logits allows for straightforward application in heterogeneous KD scenarios. Conversely, features-based distillation requires careful designs to address potential bias misalignment but has the potential for superior performance, especially on downstream tasks. In this paper, we develop a generic feature-based distillation framework and mitigate the misalignments by introducing the region-aware attention (RAA) module and adaptive feedback prompt (AFP).

\subsection{Region-Aware Attention (RAA) Module}

A key issue in heterogeneous knowledge distillation is the difference in perspective between the teacher and student models, due to their distinct architectural receptive fields. For example, a ViT-based model begins with a broad view, whereas a CNN-based model builds up to a global understanding from local views, reaching full integration in the final layers. When trying to use the traditional stage-wise feature alignment method, this difference in perspective creates a significant challenge for the student model, hindering its ability to learn effectively from the teacher model.

To address this issue, we introduce a region-aware attention (RAA) module. This module leverages an attention mechanism to help the student model learn to integrate features from different regions across stages, thereby achieving a perspective comparable to the teacher model. Owing to the flexibility of the attention module, RAA can seamlessly integrate into any architecture without requiring a distinct design. The proposed RAA is shown in Fig.~\ref{fig_framework}(a) and is composed of three steps. First, we patchify the student features from each stage $i$ to $\frac{N_q}{4}$ patches to capture regional details within each stage, and then use convolution projectors $C_i$ to map them to a predefined dimension $\mathbb{R}^{\frac{N_q}{4} \times d}$, where $N_q$ is a hyperparameter that denotes the total number of queries across all four stages. These stage features are then concatenated as $F^{S} \in \mathbb{R}^{N_q \times d}$ as depicted in Eq.~\ref{equ_FS}.
\begin{equation}
\label{equ_FS}
    F^{S} = Concat(C_1(F^{S}_1), .., C_4(F^{S}_4)),
\end{equation}
Second, We feed $F^{S}$ into a self-attention module as expressed in Eq.~\ref{equ_FS'}, which is defined as:
\begin{equation}
\label{equ_FS'}
    F^{S'} = Softmax({{(W_qF^{S})(W_kF^{S})^T}\over{\sqrt{d}}})W_vF^{S},
\end{equation}
where $W_q$, $W_k$, $W_v$ are the learnable weights, and $d$ is the channel dimension. $F^{S}$ contains features from different patches and different stages. Through the attention mechanism, the student model learns how to focus on different spatial locations from different stages to blend a new feature with a perspective similar to that of the teacher model. Finally, we divide the blended student features $F^{S'}$ back into four stages after RAA and project each to the same size as the teacher. As the blended student features approximate the teacher's perspective, standard feature matching can subsequently be applied as described in Eq.~\ref{equ_FD}.

\subsection{Adaptive Feedback Prompts (AFP) for Teacher}



The existing KD paradigm typically involves training the teacher model in isolation. This independent training means that the teacher model does not account for the subsequent distillation process intended to train the student model. Consequently, the intermediate information generated by the teacher model may not be effective for distillation, leading to suboptimal performance. This problem is particularly pronounced in heterogeneous KD, as models with different architectures have varying inductive biases, making their features less compatible with the student model.

A naive solution is to optimize the teacher model during the distillation process simultaneously \cite{guo2020onlinekd}. However, this method requires more computational cost and may lead to catastrophic forgetting \cite{kemker2018catastrophic} in the teacher model. To address this problem, we propose an adaptive feedback prompts (AFP) module, which uses prompt tuning \cite{jia2022vpt, nie2023promptTuning} to fine-tune the teacher model's knowledge with minimal parameters, aiming not only to eliminate information that is not conducive to distillation but also to make adjustments based on student feedback, while preserving the discriminative capacity of the teacher model. 

In practice, the AFP modules are inserted before the start of each stage in the teacher model, and each consists of two blocks, prompt block $PB$, and fusion block $FB$, as shown in Fig.~\ref{fig_framework}(b). To accommodate various architectures, we follow \cite{nie2023promptTuning} to develop $PB$ that integrates task-specific knowledge through an additional branch without the need to modify any architecture. Unlike the previous method, which directly feeds the teacher's features into the $PB$ \cite{zhao2023cross}, we introduce $FB$ before the $PB$ to incorporate the difference of the feature map between the student and teacher model, referred to as student feedback, into the teacher model's features, aiming to make the output prompt able to reflect the student feedback. Specifically, let $M^{AFP}_{i}$ denotes the transformation that transfers the student feature to the same dimension as the teacher's at the $i$-th stage, we subtract the student features from the teacher features in each stage to create feedback as follows:
\begin{equation}
    Feedback_{i} = M^{AFP}_{i}(F^{S}_{prev, i}) - F^{T}_{prev, i}.
\end{equation}
This feedback identifies the locations and intensities where the teacher and student features are currently inconsistent.

During forward, the AFP block inserted before $i$-th stage takes the $i$-th stage teacher feature $F^{T}_{prev, i}$ and student feature $F^{S}_{prev, i}$ from the previous iteration, along with the current $i$-th stage teacher feature $F^{T}_{i}$ as input, and outputs the final feedback enhanced teacher feature $F^{T'}_{i}$ with formula as follows: 
\begin{equation}
    F^{T'}_{i} = PB_{i}(FB_{i}(Concat(Feedback_{i}, F^{T}_{i})).
\end{equation}
We first concatenate the feedback with the teacher features and input them into a fusion block, resulting in a teacher feature that the feedback has enhanced. This feedback-enhanced feature is then sent to the prompt block to modify the teacher model's output features. In this way, the teacher model can consider the student's learning process via feedback, thus providing more friendly features to the student model during distillation. Note that the feedback is derived by subtracting the teacher and student features from the previous iteration, as utilizing the current teacher and student features to generate feedback may lead to the model easily adapting the teacher features based on this spatial-aligned feedback. Employing features from previous instances enables the model to understand the error patterns, thereby facilitating a beneficial adjustment for the student model.

Empirically, we find that the teacher features will be altered too much by the AFP modules in some teacher-student model pairs, even making the altered teacher features identical to the student's, rendering feature matching useless. To mitigate this issue, we introduce a regularization loss $L_{Reg}$ to avoid the teacher's features from converging to the student's features, thus preventing the generation of uninformative identity features by the teacher model that offers no value to the student model. The loss is defined as follows:
\begin{equation}
    L_{Reg} = L_{KL}(p^{T}, p^{T'}),
\end{equation}
where $p^{T'}$ is the predictive distribution of the teacher model incorporating the AFP modules. By ensuring that the KL-diverge distance between the predictive distributions with and without AFP modules remains minimal, we maintain the discriminative capacity of the teacher model, thereby averting the emergence of non-informative features.


\subsection{Overall Objective Function}

By integrating the proposed modules into the general knowledge distillation framework, the overall loss function of PAT is defined as: 
\begin{equation}
\label{equ_overallLoss}
    L_{PAT} = L_{CE} + \alpha L_{KL} + \beta L_{FD} + \gamma L_{Reg},
\end{equation}

where $L_{CE}$ is the standard cross-entropy loss, $L_{KD}$ denotes the KL-divergence loss between $p^T$ and $p^S$, and $L_{FD}$ denotes the feature matching loss between $F^{T'}$ and $F^{S'}$. The weights $\alpha$, $\beta$, and $\gamma$ balance the contribution of each term. For $L_{FD}$, we adopt the hierarchical context loss (HCL) from ReviewKD \cite{chen2021reviewkd} as the distance function $D$, providing stronger hierarchical supervision.

%% file: sec/4_exper.tex
\section{Experiments}
\label{sec:exper}

\definecolor{Gray}{gray}{0.85}
\newcolumntype{a}{>{\columncolor{Gray}}c}

\begin{table*}[h]
    \centering
    \resizebox{\textwidth}{!}{
    \begin{tabular}{cc|cc|cccc|cccca}
        \toprule
        \multirow{2}{*}{Teacher} & \multirow{2}{*}{Student} & \multicolumn{2}{|c}{From Scratch} & \multicolumn{4}{|c}{Logits-based} & \multicolumn{5}{|c}{Features-based} \\
        \cmidrule(r){3-13} 
         &  & T. & S. & KD & DKD & DIST & OFA & FitNet & CC & RKD & CRD & PAT \\
        \midrule
        \multicolumn{13}{c}{\textit{CNN-based students}} \\
        \midrule
        Swin-T & ResNet18 & 89.26 & 74.01 & 78.74 & 80.26 & 77.75 & \underline{80.54} & 78.87 & 74.19 & 74.11 & 77.63 & \textbf{81.22} \\
        ViT-S & ResNet18 & 92.04 & 74.01 & 77.26 & 78.10 & 76.49 & \textbf{80.15} & 77.71 & 74.26 & 73.72 & 76.60 & \underline{80.11} \\
        Mixer-B/16 & ResNet18 & 87.29 & 74.01 & 77.79 & 78.67 & 76.36 & \underline{79.39} & 77.15 & 74.26 & 73.75 & 76.42 & \textbf{80.07} \\
        Swin-T & MobileNetV2 & 89.26 & 73.68 & 74.68 & 71.07 & 72.89 & \textbf{80.98} & 74.28 & 71.19 & 69.00 & \underline{79.80} & 78.78 \\
        ViT-S & MobileNetV2 & 92.04 & 73.68 & 72.77 & 69.80 & 72.54 & \underline{78.45} & 73.54 & 70.67 & 68.46 & 78.14 & \textbf{78.87} \\
        Mixer-B/16 & MobileNetV2 & 87.29 & 73.68 & 73.33 & 70.20 & 73.26 & \textbf{78.78} & 73.78 & 70.73 & 68.95 & 78.15 & \underline{78.62} \\
        \midrule
        \multicolumn{13}{c}{\textit{ViT-based students}} \\
        \midrule
        ConvNeXt-T & DeiT-T & 88.41 & 68.00 & 72.99 & 74.60 & 73.55 & \underline{75.76} & 60.78 & 68.01 & 69.79 & 65.94 & \textbf{79.59} \\
        Mixer-B/16 & DeiT-T & 87.29 & 68.00 & 71.36 & 73.44 & 71.67 & \underline{73.90} & 71.05 & 68.13 & 69.89 & 65.35 & \textbf{74.66} \\
        ConvNeXt-T & Swin-P & 88.41 & 72.63 & 76.44 & 76.80 & 76.41 & \underline{78.32} & 24.06 & 72.63 & 71.73 & 67.09 & \textbf{80.74} \\
        Mixer-B/16 & Swin-P & 87.29 & 72.63 & 75.93 & 76.39 & 75.85 & \textbf{78.93} & 75.20 & 73.32 & 70.82 & 67.03 & \underline{78.44} \\
        \midrule
        \multicolumn{13}{c}{\textit{MLP-based students}} \\
        \midrule
        ConvNeXt-T & ResMLP-S12 & 88.41 & 66.56 & 72.25 & 73.22 & 71.93 & \underline{81.22} & 45.47 & 67.70 & 65.82 & 63.35 & \textbf{83.50} \\
        Swin-T & ResMLP-S12 & 89.26 & 66.56 & 71.89 & 72.82 & 11.05 & \underline{80.63} & 63.12 & 68.37 & 64.66 & 61.72 & \textbf{80.94} \\
        \midrule
        \midrule
         \multicolumn{4}{c|}{\textit{Average Improvement}} & 3.17 & 3.16 & -2.31 & \underline{7.47} & -5.20 & -0.33 & -1.40 & -0.02 & \textbf{8.17} \\
        
        \bottomrule
    \end{tabular}
    }
    \caption{Result on CIFAR-100. Our results are the average over 3 trials. The highest results are indicated in bold, while the second-best results are underlined.}
    \label{tab:cifar}
\end{table*}

\begin{table*}[t]
    \centering
    \resizebox{\textwidth}{!}{
    \begin{tabular}{cc|cc|cccc|cccca}
        \toprule
        \multirow{2}{*}{Teacher} & \multirow{2}{*}{Student} & \multicolumn{2}{|c}{From Scratch} & \multicolumn{4}{|c}{Logits-based} & \multicolumn{5}{|c}{Features-based} \\
        \cmidrule(r){3-13} 
         &  & T. & S. & KD & DKD & DIST & OFA & FitNet & CC & RKD & CRD & PAT \\
        \midrule
        \multicolumn{13}{c}{\textit{CNN-based students}} \\
        \midrule
        Swin-T & ResNet18 & 81.38 & 69.75 & 71.14 & 71.10 & 70.91 & \textbf{71.85} & 71.18 & 70.07 & 68.89 & 69.09 & \underline{71.54} \\
        Mixer-B/16 & MobileNetV2 & 76.62 & 68.87 & 71.92 & 70.93 & 71.74 & \underline{72.12} & 71.59 & 70.79 & 69.86 & 68.89 & \textbf{72.22} \\
        \midrule
        \multicolumn{13}{c}{\textit{ViT-based students}} \\
        \midrule
        ConvNeXt-T & DeiT-T & 82.05 & 72.17 & 74.00 & 73.95 & 74.07 & \underline{74.41} & 70.45 & 73.12 & 71.47 & 69.18 & \textbf{74.44} \\
        \midrule
        \multicolumn{13}{c}{\textit{MLP-based students}} \\
        \midrule
        Swin-T & ResMLP-S12 & 81.38 & 76.65 & 76.67 & 76.99 & 77.25 & \underline{77.31} & 76.48 & 76.15 & 75.10 & 73.40 & \textbf{77.59}\\
        \midrule
        \midrule
         \multicolumn{4}{c|}{\textit{Average Improvement}}  & 1.57 & 1.38 & 1.63 & \underline{2.06} & 0.57 & 0.67 & -0.53 & -1.72 & \textbf{2.09} \\
        \bottomrule
    \end{tabular}
    }
    \caption{Result on ImageNet. Our results are the average over 3 trials. The highest results are indicated in bold, while the second-best results are underlined.}
    \label{tab:imagenet}
    \vspace{-10pt}
\end{table*}

\begin{table}[]
    \centering
    \resizebox{\linewidth}{!}{
    \begin{tabular}{c|ccc|ccc}
    \toprule
     & \multicolumn{3}{c|}{Swin-T \& ResNet18} & \multicolumn{3}{c}{Swin-T - MobileNetV2} \\
     & mAP & AP50 & AP75 & mAP & AP50 & AP75 \\
    \midrule
     Teacher & 45.14 & 67.09 & 49.25 & 45.14 & 67.09 & 49.25 \\
     Student & 33.26 & 53.61 & 35.26 & 29.47 & 48.87 & 30.90 \\
    \midrule
    KD      & 34.07 & 55.26 & 36.48 & 31.46 & 52.40 & 32.74 \\
    DKD     & 29.96 & 51.17 & 31.36 & 32.10 & \underline{53.82} & 33.88 \\
    OFA     & 33.37 & 54.98 & 35.13 & 31.69 & 52.91 & 32.88 \\
    \midrule
    FitNet      & \underline{35.23} & \underline{56.09} & \underline{37.31} & \underline{32.48} & 52.62 & \underline{34.67} \\
    \rowcolor{Gray}
    PAT         & \textbf{35.62} & \textbf{56.67} & \textbf{38.04} & \textbf{32.97} & \textbf{54.18} & \textbf{35.08} \\
    \bottomrule
    \end{tabular}
    }
    \caption{Results on COCO based on Faster RCNN \cite{ren2015fasterrcnn} with FPN \cite{lin2017pyramid}. The highest results are indicated in bold, while the second-best results are underlined.}
    \label{tab:coco}
    \vspace{-10pt}
\end{table}

We conduct experiments on various tasks. First, we compare PAT with other KD methods on two image classification datasets. We then apply our method to object detection to show the pros of feature-based distillation. All results of PAT are reported by taking the average over 3 trials. Due to the space constraint, please refer to the supplementary for the implementation details.


\noindent
\textbf{Datasets} \hspace{5pt} Two image classification datasets and one detection dataset are adopted for our experiments. (1) CIFAR-100 \cite{krizhevsky2009learning} is a well-known dataset, containing 50K training and 10K test images of size 32×32 in 100 classes. (2) ImageNet \cite{deng2009imagenet} is a large-scale classification dataset, which provides 1.2M images for training and 50K images for validation over 1,000 classes. (3) COCO \cite{cocodataset} is an 80-category object detection dataset, which contains 118K images for training and 5K images for validation.

\vspace{5pt}

\noindent
\textbf{Models} \hspace{5pt} We evaluate our PAT on various networks, including ResNet \cite{kaiming2016resnet}, MobileNetv2 \cite{sandler2018mobilenetv2}, and ConvNeXt \cite{liu2022convnext} as CNN-based models, ViT \cite{dosovitskiy2021an}, DeiT \cite{touvron21a} and Swin Transformer \cite{liu2021swin} as ViT-based models, and MLP-Mixer \cite{tolstikhin2021mlpmixer} and ResMLP \cite{touvron2023resmlp} as MLP-based models. 

\vspace{5pt}

\noindent
\textbf{Baselines} \hspace{5pt} We compare with various approaches that are feature-based, such as FitNet \cite{adriana2015fitnet}, CC \cite{peng2019cc}, RKD \cite{park2019rkd}, and CRD \cite{tian2019crd}, as well as the baselines that are based on logits, which include KD \cite{Hinton2015DistillingTK}, DKD \cite{zhao2022dkd}, DIST \cite{huang2022knowledge} and the most recent state-of-the-art (SOTA) OFA-KD \cite{hao2023ofa}.




\subsection{Results on Image Classification}



On CIFAR-100, we compare our PAT method with eight other KD methods, encompassing a total of twelve heterogeneous teacher-student pairings. As shown in Table~\ref{tab:cifar}, PAT outperforms other methods on most teacher-student pairs, resulting in an average improvement of 8.17\%. For instance, we achieve an accuracy of 83.50\% on the ConvNeXt-T - ResMLP-S12 pair and 79.59\% on the ConvNeXt-T - DeiT-T pair, surpassing not only the other feature-based methods but also obtaining an improvement of 2.28\% and 3.83\% compared to the prior SOTA logits-based method OFA. Specifically, PAT does not achieve as much improvement on CNN-based students as it does on ViT-based and MLP-based students. Unlike the other two types of students, CNN-based students need additional patch-embedding layers, we speculate that these additional parameters could result in prolonged training epochs, thus only having a small improvement in current settings.


Due to computational limitations, we select four teacher-student settings to experiment with on ImageNet. The results are presented in Table~\ref{tab:imagenet}. With an increase in the volume of training data, the majority of methods exhibit a favorable average enhancement. Among them, PAT attains SOTA performance across most teacher-student pairs. For instance, we achieve 72.22\% on MobileNetV2 and 77.59\% on ResMLP-S12 when distilling from the teacher model Mixer-B/16 and Swin-T, which is 3.35\% and 0.94\% higher than the baselines, proving our generalizability on a larger dataset.


Previous feature-based methods achieve inferior performance compared to logits-based methods, even getting a negative average improvement. Methods like CC, RKD, and CRD learn from the relationships between the data samples rather than directly imitating features, making them less vulnerable to view mismatch. However, they still underperform as they ignore the teacher unawareness problem. FitNet and our method both imitate the teacher model's features, however, without considering the view mismatch problem, the imitation is less efficient and thus performs badly. Our performance suggests that by tackling the discrepancy in perspectives and lack of awareness in the instructor, a feature-imitated approach can achieve comparable or even superior performance in heterogeneous KD scenarios.





\subsection{Results on Object Detection}

We extend our method to object detection and follow the setting of DKD \cite{zhao2022dkd} to perform distillation between the teacher's and student's backbone features. The results are presented in Table~\ref{tab:coco}. Logits-based methods, as mentioned in DKD, can only achieve inferior performance as they disregard the intermediate features. For example, OFA can not reach good performance owing to the excessive focus on semantic details instead of spatial aspects, which is crucial for detection. Conversely, feature-based methods demonstrate superior performance, with our approach achieving SOTA results. We observe an enhancement of 2.36\% for ResNet18 and 3.50\% for MobileNetV2 when distilling from the Swin-T teacher model. This underscores that by mitigating view mismatch and teacher unawareness issues, the feature-mimicking technique can effectively leverage the abundant intermediate features for improved performance across classification and downstream tasks.




\begin{table}[hbt!]
    \centering
    \begin{tabular}{cccc}
        \toprule
        RAA & AFP w/o Feedback & AFP & Acc. \\
        \midrule
        \multicolumn{3}{c}{Baseline (FitNet)}   & 60.71 \\
        \midrule
        \checkmark &   &   & 70.12 \\
        \checkmark & \checkmark &   & 79.13 \\
        \checkmark &   & \checkmark & 79.59 \\
        \bottomrule
    \end{tabular}
    \caption{The effectiveness of our proposed RAA, AFP on CIFAR-100 with ConvNeXt-T - DeiT-T as the teacher-student pair.}
    \label{tab:ablation_ProposedModule}
    \vspace{-10pt}
\end{table}

%% file: sec/5_ablation.tex
\section{Ablation Studies}
\label{sec:ablation}

\subsection{Effectiveness of each proposed module}

Experiments are conducted to evaluate the effectiveness, where the proposed modules are gradually integrated to measure their influence, and the results are summarized in Table~\ref{tab:ablation_ProposedModule}. By leveraging the RAA module, the issue of view mismatch is resolved, leading to a significant improvement in performance compared to the baseline. When we further introduce the AFP module to adapt the teacher model, the student yields larger gains, and the performance becomes the best when student feedback is incorporated into the AFP.

\subsection{Effectiveness of number and position of AFP}
\label{section_effectiveOfNumberAndPositionOfAFP}


\begin{table}[]
    \centering
    \resizebox{\linewidth}{!}{
    \begin{tabular}{cccc|cc}
        \toprule
        Stage 1 & Stage 2 & Stage 3 & Stage 4 & Mem. (GB) & Acc. \\
        \midrule
        \multicolumn{4}{c|}{Baseline (FitNet)} & 6.12 & 60.71 \\
        \midrule
         \checkmark &   &   &   & 10.76 & 75.43 \\
           & \checkmark &   &   & 11.55 & 77.34 \\
           &   & \checkmark &   & 10.30 & 76.26 \\
           &   &   & \checkmark &  9.01 & 69.65 \\
        \midrule
         \checkmark & \checkmark & \checkmark &   & 13.70 & 77.87 \\
           & \checkmark & \checkmark & \checkmark & 12.31 & 78.71 \\
         \checkmark & \checkmark & \checkmark & \checkmark & 14.02 & 79.59 \\
        \bottomrule
    \end{tabular}
    }
    \caption{Effect of number and position of AFP on CIFAR-100 with ConvNeXt-T - DeiT-T as the teacher-student pair.}
    \label{tab:ablation_AFPStage}
\end{table}

Since adopting AFP at every stage is beneficial yet introduces additional parameters, we conduct experiments to investigate the effectiveness of adopting AFP at each stage as in Table ~\ref{tab:ablation_AFPStage}. Enhancements in performance are observed even with the adoption of AFP in just one of the stages, with optimal outcomes achieved when AFP is implemented across all stages but also with the highest memory usage. In our experiments, we adopt AFP at each stage of the teacher model by default to achieve the best performance. 

\subsection{Number of queries on RAA}
\label{section_numberOfQueriesOnRAA}

\begin{table}[]
    \centering
    \begin{tabular}{ccc}
        \toprule
        $N_q$ & Mem. (GB) & Acc. \\
        \midrule
        Baseline (FitNet) & 6.12 & 60.71 \\
        \midrule
        36  & 13.79 & 79.03 \\ 
        64  & 14.02 & 79.59 \\
        80  & 14.37 & 80.14 \\
        144 & 14.54 & 80.89 \\
        \bottomrule
    \end{tabular}
    \caption{Comparation of the $N_q$ in RAA on CIFRA-100. We adopt ConvNeXt-T - DeiT-T as the teacher-student pair.}
    \label{tab:ablation_RAANQuery}
    \vspace{-5pt}
\end{table}

In RAA, each stage feature of the student is projected into a predefined sequence of shape $\mathbb{R}^{\frac{N_q}{4} \times d}$ for attention computation. Increasing $N_q$ allows finer subdivision of features, offering more detailed representations, but also leads to larger attention matrices and higher computational cost. We evaluate the impact of varying $N_q$ from 36 to 144 on performance and memory usage. As shown in Table~\ref{tab:ablation_RAANQuery}, performance improves steadily with larger $N_q$, accompanied by proportional memory growth. Notably, setting $N_q=64$ achieves SOTA on CIFAR-100 with minimal memory usage. By default, we use $N_q=64$ for CIFAR-100 and $N_q=144$ for larger datasets such as ImageNet and COCO.


\subsection{The performance of PAT without KL loss}

\begin{table}[]
    \centering
    \resizebox{\linewidth}{!}{%
    \begin{tabular}{cccc}
        \toprule
        Teacher & Swin-T & ConvNeXt-T & ConvNeXt-T \\
        Student & ResNet18 & DeiT-T & ResMLP-S12 \\
        \midrule
        KD               & 78.74 & 72.99 & 72.25 \\
        FitNet           & 78.87 & 60.78 & 45.47 \\
        PAT w/o KL Loss & 79.14 & 75.98 & 81.79 \\
        PAT w/  KL Loss & 81.22 & 79.10 & 83.50 \\ 
        \bottomrule
    \end{tabular}
    }
    \caption{The effectiveness of loss $L_{KL}$ on PAT on CIFAR-100.}
    \label{tab:fofa_wo_kdloss}
    \vspace{-5pt}
\end{table}


PAT is not purely feature-based, as it also incorporates a logits-based KL-divergence loss, as shown in Eq.\ref{equ_overallLoss}. This raises the question of whether performance gains stem primarily from the logits-matching loss rather than the feature-level alignment. To investigate, we train PAT without the KL loss. As shown in Table\ref{tab:fofa_wo_kdloss}, adding the KL loss improves performance. However, even without it, PAT still achieves performance close to SOTA. For best results, we include the KL loss by default to further enhance overall effectiveness.

\subsection{Complexity Analysis}

\begin{table}[]
    \centering
    \begin{tabular}{cccc}
    \toprule
    Method & Extra Para. (M) & Epoch (s) & Acc.   \\ 
    \midrule
    KD     &   0   & 65.75 & 72.99 \\
    OFA    &  3.78 & 76.16 & 75.76 \\
    \midrule
    FitNet &  1.11 &  90.45 & 60.78 \\
    PAT   & 14.48 & 208.57 & 79.59 \\
    \bottomrule
    \end{tabular}%
    \caption{The complexity of KD methods with ConvNeXt-T - DeiT-T as the teacher-student pair on CIFAR-100.}
    \label{tab:table1}
    \vspace{-10pt}
\end{table}

We also explore the complexity of our method. As shown in Table~\ref{tab:table1}, it is found that logits-based methods, which rely solely on logits for distillation, require significantly fewer parameters compared to feature-based methods, leading to faster training times. In contrast, feature-based methods involve more parameters, which extends the training duration. Although PAT consumes the most parameters, it also achieves the highest performance. Notably, these additional parameters only appear during training and do not add any overhead to the student model during inference. 


For the trade-off between the complexity and performance of PAT, as mentioned in section ~\ref{section_effectiveOfNumberAndPositionOfAFP} and ~\ref{section_numberOfQueriesOnRAA}, we can achieve faster training by lowering the number of AFP modules, or by decreasing the queries number on RAA.



\subsection{Visualization of attention map within RAA}

\begin{figure}
    \centering
    \includegraphics[width=1.0\linewidth]{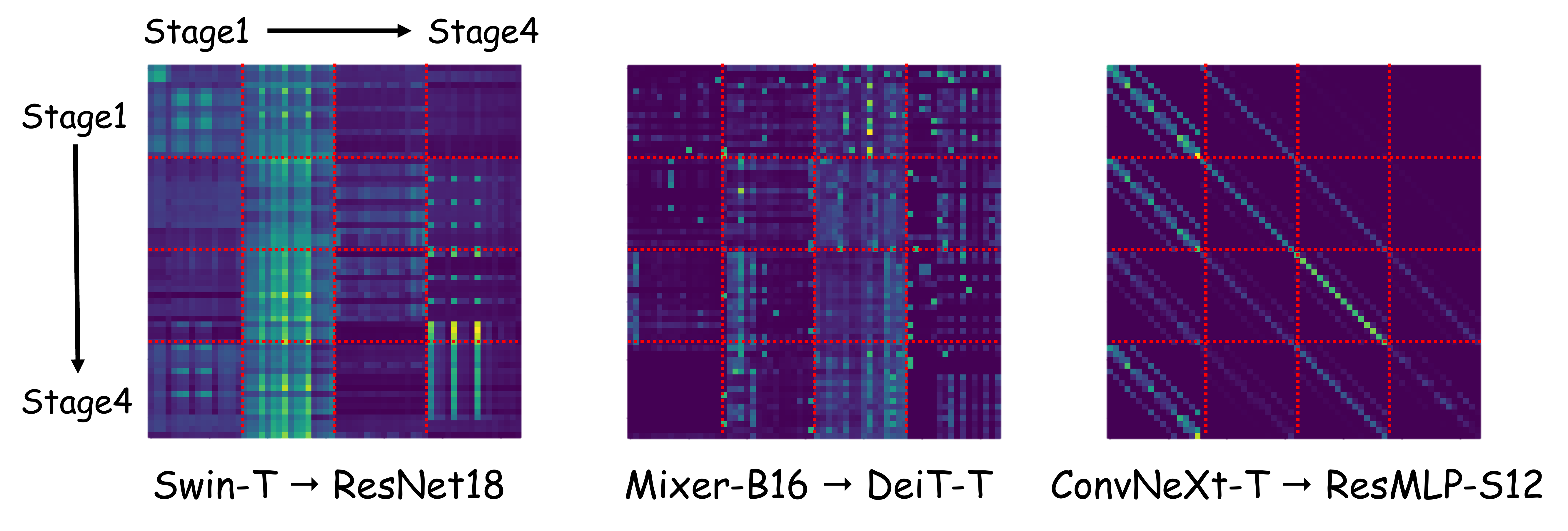}
    \caption{Visulizations of the attention map within RAA from different model pairs. Queries are sorted based on patch and stage position, as the black arrow shows.}
    \label{fig:visualization_raa}
    \vspace{-10pt}
\end{figure}


Direct feature matching in heterogeneous KD is suboptimal due to differing inductive biases across architectures, leading to mismatched feature representations. To address this, PAT leverages attention to guide the student in blending features from different regions and stages, aligning its perspective with the teacher. As shown in Fig.~\ref{fig:visualization_raa}, attention patterns vary across teacher-student pairs. For example, the MLP-based student ResMLP-S12 learns from the CNN-based teacher ConvNeXt-T by attending to consistent spatial positions across stages, forming diagonal patterns that reflect local aggregation. In contrast, the CNN-based student ResNet18, learning from the ViT-based teacher Swin-T, captures features from neighboring patches, resulting in grid-like patterns indicative of global context aggregation. Similarly, the ViT-based student DeiT-T, when guided by the MLP-based teacher Mixer-B16, exhibits global attention with a strong focus on stage 3, suggesting that key information is concentrated there.

%% file: sec/6_conclusion.tex
\section{Conclusion}
\label{sec:conclusion}

In this paper, we propose a perspective-aware teaching (PAT) framework to facilitate feature-level supervision in the context of heterogeneous KD. First, we introduce region-aware attention to blend the stage features of the student model to align with those of the teacher model, thereby alleviating the issue of view mismatch prevalent between the two models. Moreover, we incorporate the prompt tuning technique to address the teacher unawareness problem and enable to adapt the teacher model's feature utilizing feedback from the student model with minimal parameters. Through the integration of these two components, we can realize feature-level distillation in heterogeneous KD scenarios and achieve promising performance in classification and downstream tasks. 


\section{Acknowledgments}
This work is partially supported by the National Science and Technology Council, Taiwan, under Grant: NSTC-114-2640-E-A49-011, NSTC-114-2218-E-A49-017, and NSTC-112-2628-E-002-033-MY4, and was financially supported in part by the Center of Data Intelligence: Technologies, Applications, and Systems, National Taiwan University (Grants: 114L900901/114L900902/114L900903), from the Featured Areas Research Center Program within the framework of the Higher Education Sprout Project by the Ministry of Education, Taiwan.
